\documentclass{article}
\usepackage{spconf,amsmath,epsfig}
\usepackage{amssymb}
\usepackage{graphicx}
\usepackage{mathtools}
\usepackage {diagbox}
\usepackage{makecell}
\usepackage{wasysym}
\usepackage{hyperref}
\usepackage{comment}
\usepackage{enumerate}
\usepackage{setspace}
\usepackage{booktabs}
\let\OLDthebibliography\thebibliography
\renewcommand\thebibliography[1]{
  \OLDthebibliography{#1}
  \setlength{\parskip}{0pt}
  \setlength{\itemsep}{0pt plus 0.3ex}
}

\begin{document}\sloppy

\def\x{{\mathbf x}}
\def\L{{\cal L}}

\title{Pyramidal Hidden Markov Model For Multivariate Time Series Forecasting}
%
\name{YeXin Huang, Independent Researcher}
\address{}

\maketitle

\begin{abstract}
\frenchspacing

The Hidden Markov Model (HMM) can predict the future value of a time series based on its current and previous values, making it a powerful algorithm for handling various types of time series. 
Numerous studies have explored the improvement of HMM using advanced techniques, leading to the development of several variations of HMM. 
Despite these studies indicating the increased competitiveness of HMM compared to other advanced algorithms, few have recognized the significance and impact of incorporating multistep stochastic states into its performance. 
In this work, we propose a Pyramidal Hidden Markov Model (PHMM) that can capture multiple multistep stochastic states. 
Initially, a multistep HMM is designed for extracting short multistep stochastic states. 
Next, a novel time series forecasting structure is proposed based on PHMM, which utilizes pyramid-like stacking to adaptively identify long multistep stochastic states. 
By employing these two schemes, our model can effectively handle non-stationary and noisy data, while also establishing long-term dependencies for more accurate and comprehensive forecasting.
The experimental results on diverse multivariate time series datasets convincingly demonstrate the superior performance of our proposed PHMM compared to its competitive peers in time series forecasting. 
\end{abstract}
\section{Introduction}
\label{sec:intro}
\frenchspacing
Time series forecasting is crucial in various social domains, facilitating resource management and decision-making.
For instance, in medical diagnosis, past data assist in the timely detection and control of potential diseases. 
Early forecasting usually presents some challenges, including:
i) violating stability assumptions of traditional models hinders accurate prediction of future trends, leading to non-stationarity \cite{NEURIPS2022_4054556f};
and ii) undergoing distributional shifts across various working conditions results in Out-Of-Distribution (OOD), which poses a challenge for many existing supervised learning models \cite{yang2021oodsurvey}.
Hidden Markov Models (HMM) are frequently employed to tackle nonstationary data and OOD forecasting due to their ability to operate with stochastic states rather than fitting a single assumed distribution \cite{chatzis2012reservoir, sun1996non, jinghui2005principles}.

However, most of existing HMMs, only accounting for single-step stochastic state transitions, encounters long-term forecasting challenges when applied to real-world time series that can be described as predominantly consisting of multiple multistep stochastic state.
Consequently, such a conventional HMM will be hard to achieve forecasting well, especially in terms of long-term prediction.
Thus, in this work, to enhance the identification of multistep stochastic states, we introduce a novel Hidden Markov Model with multiple steps. 
Our model incorporates an attention mechanism \cite{vaswani2017attention} into the data input of a multistep time window, enabling realistic capture of these states in a time series.
Moreover, we employ a stacking technique to appropriately increase the depth of the model structure, enhancing its adaptability to stochastic states with varying step sizes and improving its predictive capability through deep stacking.

\begin{figure}[t]
  \centering 
  \includegraphics[width=8cm]{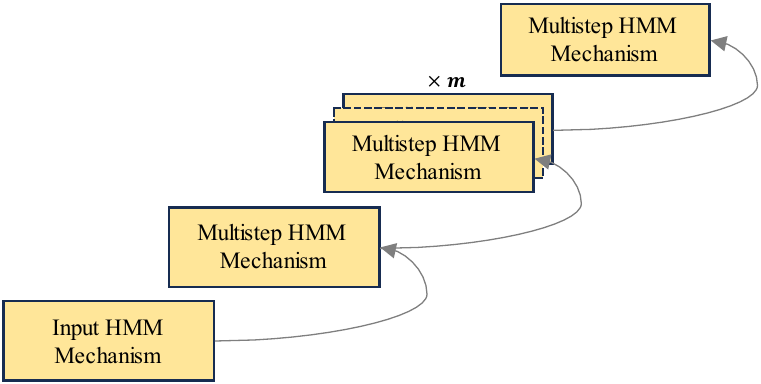}
  \caption{Brief structure of PHMM. By stacking these Multistep HMM Mechanisms on the Input HMM mechanism, we are able to achieve a higher accuracy and efficiency in various applications. The additional layers provide a deeper understanding and analysis of complex data patterns, allowing for more precise predictions and decision-making.}
  \label{fig:1}
  \vspace{-0.5cm}
\end{figure}
\begin{figure*}[ht]
  \centering
  \includegraphics[width=17cm]{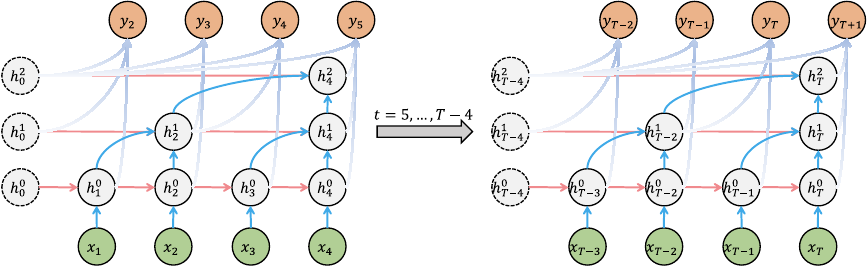}
  \caption{Directed Acyclic Graphs for Pyramidal Hidden Markov Model (where $k=2, m=3$) .
  The variable $k$ denotes the time step, the variable $m$ denotes the number of stacked layers and the latent variable $h_{t}^{i}(i=1, 2, \dots, m)$ denotes the latent variable of different level.
  The hidden variables independently propagate as $t$ grows. 
  At each step, the $x_{t-1}$ generate the outputs $y_t$'s lowest latent variable, and subsequently, this lower-level latent variable is propagated to the higher level in order to generate the longer stochastic state.
  Note that gray variable means unobserved variable, other colored variable means observed one.}
  \label{fig:DAG}
\end{figure*}
We utilize a training method based on deep generative models called a Variational Autoencoder (VAE) , which learn from a low-dimensional potential space assumed to reside on known Riemannian manifolds \cite{kingma2022autoencoding}. 
Moreover, the neutralized HMM is optimized using the optimizing the Evidence Lower Bound (ELBO) and the reparameterization trick \cite{Mehta_2022}, as opposed to the traditional Expectation-Maximization (EM) method. 
Then, the proposed model is trained on both small and large datasets without imposing strong regularization.
The experimental results provide convincing evidence that the proposed PHMM outperforms its competitors for time series forecasting and effectively addresses the limitations of the conventional HMM in capturing the multistep stochastic state, as shown in Table \ref{tab:1}.
The ablative studies demonstrate the influence of varying timesteps and stacked layer numbers in our method, thereby providing compelling evidence to substantiate the efficacy of PHHM.
Our contribution can be summarized as follows:
\begin{enumerate}[0]
  \item [$\bullet$] \textbf{Methodologically}, we propose a novel Pyramidal Hidden Markov Model of sequential data for future forecasting;
  \item [$\bullet$] \textbf{Algorithmically}, we reformulate a new sequential VAE framework, which is aligned with the proposed model above;
  \item [$\bullet$] \textbf{Experimentally}, we achieve SOTA prediction result on 20 UEA datasets. 
  In our ablation study, we also demonstrate the efficiency of PHHM in comparison to conventional HMM structure.
\end{enumerate}
\begin{table}[ht]
  \centering
  \vspace{-0.5cm}
  \caption{Others VS. Our model.}
  \label{tab:1}
  \resizebox{\columnwidth}{!}{
    \begin{tabular}{c|c c c c c c|c}
      \hline
      Method           & \makecell{WEASEL\\+MUSE} & \makecell{MLSTM\\-FCN} & \makecell{TapNet} & \makecell{ShapeNet} & \makecell{ROCKET}& \makecell{RLPAM} & Wins \\
      \hline
      WEASEL+MUSE    &                                  & \checked                            &           &               &         &   &  7  \\
      MLSTM-FCN     & \checked                                &                              &           &               &         &   & 9   \\
      TapNet        & \checked                                & \checked                            &             &              &   &         &    7\\
      ShapeNet       & \checked                                & \checked                            & \checked           &          &      &            & 7   \\
      ROCKET       &                                 &                             & \checked           &               &         &     & 8    \\
      RLPAM       & \checked                                & \checked                            & \checked           &  \checked        & \checked    &     & 10   \\
      \hline
      Our Model     & \checked    & \checked                            & \checked           & \checked              & \checked         & \checked   & 12   \\
      \hline
    \end{tabular}
  }
\end{table}

\section{Related Work}
Traditional time series forecasting methods typically require continuous, complete time series data, which refers to labeled values or supervisory signals at each time step for future phases, e.g., \cite{yue2022ts2vec, wang2022learning, woo2022cost, jin2022domain}. 
However, our time series forecasting scenario can tolerate missing data or outliers. In real-world scenarios, outliers are often observed due to noise in sensor sampling or errors in data transmission \cite{xu2021anomaly, yue2022ts2vec}.

Extensive research has been conducted on the integration of neural networks with Hidden Markov Model (HMM) to maintain its unique characteristics \cite{Mehta_2022}.
As an example, researchers have successfully utilized cyclic units in the hidden semi-Markov model (HSMM) for accurate segmentation and labeling of complex time series data \cite{yu2010hidden}.
Moreover, HMM finds application in analyzing medical time series data for predicting pathologies \cite{kawamoto2013hidden, li2021causal}.
Nevertheless, many previous studies overlook the significance of incorporating multistep features to establish long-term dependencies when designing hidden Markov models.
Furthermore, the conventional model may encounter difficulties in attaining stable training or effectively converging due to the extensive duration of time series involved during the training process.
Conversely, our novel approach focuses on capturing multistep random states as a means to establish significant long-term dependencies, which is an aspect frequently disregarded by earlier investigations. 
The pyramidal structure of our proposed method make it easily trainable.

Other approaches similar to ours are modeling time series data \cite{li2021causal}. 
Most of these works use Multilayer Perception (MLP) or Convolutional Neural Networks (CNN) for feature extraction and neural networked HMMs are used to generate trajectories for extracting features \cite{gao2018brain, cui2018longitudinal}. 
However, these methods often suffer from structural flaws due to the extensive use of nonlinear activation, which can lead to a reduced predictive capability over long time series and hinder the ability to identify potential external causality.
Deep generative networks are used to learn from low-dimensional Riemannian spaces by training the encoder. An RNN is employed to encode sequential data into a hidden representation and then decode it to generate the original sequential data \cite{hewamalage2021recurrent}.
We further incorporate the VAE architecture and reformulate it, facilitating the training of both encoder and decoder components.

\section{Preliminaries}
\textbf{Directed Acyclic Graph (DAG).} The DAG is denoted as $G\in(V, E)$ with $V$, $E$ respectively denoting the node and edge set \cite{pearl2009causality}.
Each arrow $x \to y$ in $E$ represents a direct effect of $x$ on $y$.
The structural equations define the generating mechanisms for each each node in $V$. 
Specifically, for $V\in\{v_1, ...,v_k\}$ ,the mechanisms associated with the structural equations(defined  as ${f_i}_{v_i \in V}$) are defined as:$\{v_i \gets f_i(Pa(V_i),{\epsilon}_i)\}_{v_i \in V}$.

\frenchspacing
To further visualise our model, our DAG is given as shown in Fig. \ref{fig:DAG}. 
We introduce hidden variables ${h_t^1, h_t^2, \ldots, h_t^m}$ which are transmitted within these Hidden Markov Chains. 
Each hidden variable $h_t^m$ ($m=1,2,\ldots,m$) is generated from the outputs ${h_{t-k+1}^{m-1}, h_{t-k+2}^{m-1}, \ldots, h_t^{m-1}}$ and $h_{t-1}^m$. 
Finally, the hidden variables from each chain are concatenated and used as output for performing the prediction task at each step. 
Subsequently, we provide a detailed explanation of the model definition and the learning methodology in the subsequent sections.

\section{Model}
\textbf{Problem Statement.} A multivariate time series can be represented as $X=\{X_1,X_2,\ldots,X_n\}$, where $n$ denotes the number of samples. 
Each sample $X_i$ consists of a series of events, $X_i=\{x_{i_{1}},x_{i_{2}},\ldots,x_{i_{m}}\}$, where each $x_{ij}$ denotes the value of the observed sequence at each time step $t$, and each time step can encompass a different number of record dimensions. In particular, $x_{i_{j}} \in \mathbb{R}^D$.
Our goal is to train a prediction model that can perform both the classification task $f: X_i \rightarrow y_i$ and the prediction task $f: x_{i_{m}} \rightarrow x_{i_{m+1}},\ x_{i_{m+2}},\ldots,\ x_{i_{m+p}}$, where $p$ denotes the number of prediction steps.
\\
\textbf{Outline.} We first introduce our Pyramidal Hidden Markov
Model in section 4.1. Then we introduce our method to learn
these hidden variables in section 4.2.
\subsection{Pyramidal Hidden Markov Model}
\begin{figure*}[ht]
  \centering
  \includegraphics[width=\textwidth]{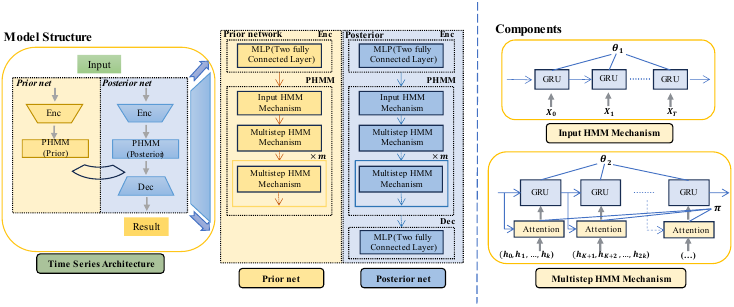}
   \caption{Left: The illustration of the time series architecture for our proposed PHMM. 
   At each time step $t$, for the prior network, it takes the concatenated features(i. features via encoder on time series input; ii. the hidden variable at the last step) into PHMM (prior). The PHMM (prior) is followed with two FC layers which output the mean and log-variance vector respectively. After sampling, the prior hidden variables are obtained. For the posterior network, the input $x_{t}$ is processed by fully connected layers to extract corresponding features. Then, the features are also process by the PHMM (posterior) which is followed by two FC layers that output the mean and log-variance vector respectively. 
   The hidden variances of Input HMM Mechanism are then fed into decoder network for regenerate the input.
   Finally, all hidden variables generated by both Input HMM Mechanism and Multistep HMM Mechanisms are concated to be fed into the predictor to predict the future values.
   Right: The structure of two basic components at each time step $t$. The Input HMM Mechanism consists of a shared GRU cell with parameter $\theta_{1}$. The Multistep HMM Mechanism consists of a shared GRU cell with parameter $\theta_{2}$ and an attention Mechanism with parameter $\pi$, which captures short-term dependencies based on $k$ hidden variables from the lower layer.}
  \label{fig3}
\end{figure*}

\frenchspacing
To describe the model design, we introduce the directed acyclic graph of DAG as shown in Fig \ref{fig:DAG}. As a new Hidden Markov Model design, our model is named Pyramidal Hidden Markov Model. the overall structure of our model is depicted as Fig \ref{fig3}. Our model is defined in detail as follows:
\\
\\
\textbf{Definition 4.1 (Pyramidal Hidden Markov Model)} Pyramidal Hidden Markov Model of directed acyclic graph DAG is formed. where the model is defined as $F$, $F \triangleq F_{t}$, where $F_{t} \triangleq x_{t} \leftarrow f(x_{t},\epsilon_{x_{t}}, h_{t}^{1}, \epsilon_{h_{t}^{1}}, h_{t}^{2}, \epsilon_{h_{t}^{2}}, \ldots, h_{t}^{m}, \epsilon_{h_{t}^{m}},)$ for $t < T$. $F_{T}$ additionally contains $y_{t} \leftarrow f(y, h_{t}^{1}, h_{t}^{2}, \ldots, h_{t}^{m})$. 
where ${\epsilon_{x_{t}}, \epsilon_{h_{t}^{1}}, \epsilon_{h_{t}^{2}}, \ldots, \epsilon_{h_{t}^{m}}}$ are independently exogenous variables.
For $t<T$, similarly, $\epsilon_{x_{t}}$,\space$\epsilon_{h_{t}^{1}}$, $\epsilon_{h_{t}^{2}}$, $\ldots$, $\epsilon_{h_{t}^{m}}$ are the latent variables at level of $1,2,...,m$ and identically distributed at time step $t$.
\\
\\
\textbf{Input Hidden Markov Mechanism}

\noindent The Input Hidden Markov Model is implemented using a neuralized HMM, represented in the Fig \ref{fig3} as a Gated Recurrent Unit (GRU) \cite{chung2014empirical} with shared parameters $\theta_{1}$ and the latent variable $h_t^1$. 
At each time step, an output result $o_t$ is generated and retained for both output and input in multistep HMM mechanisms.
The detailed definitions of prior and posterior are provided below.

Based on the Input Markov Condition \cite{pearl2009causality}, we have the following factorization of joint distribution as
\begin{eqnarray}
  &p(h_{t<T}^1,x_{t<T},y_T) = p(y_T | h_T^1)* \nonumber\\
  &\prod_{t=1}^{T-1}\Big(p(h_t^1 | h_{t-1}^1,x_{t-1})p(x_t|h_t^1) \Big),
\end{eqnarray}
$\text{The} \left\{p\left(h_t \middle| h_{t-1}^1,x_{t-1}\right)\right\}_t, q_\phi\left(h_{t<T}^1 \middle| x_{t<T},y_T\right), \left\{p(x_t \middle| h_t^1)\right\}_t$ denote the a priori, a posteriori, and a generative model respectively.

\noindent\textbf{Priori.}
For a priori $p_\psi\left(h_t^1 \middle| h_{t-1}^1,x_{t-1}\right)$, it can be written as: 
\begin{eqnarray}
  p_\psi\left(h_t^1 \middle| h_{t-1}^1,x_{t-1}\right) = \prod_{o}{p_\psi\left(o_t^1 \middle| o_{t-1}^1,x_{t-1}\right)},
\end{eqnarray}
\frenchspacing
where for any $o^1\in\left\{h^1\right\}$, $p_\psi\left(h_t^1 \middle| h_{t-1}^1,x_{t-1}\right)$ for each t is distributed as $N(\mu_\psi\left(h_{t-1}^1,x_{t-1}\right), \Sigma _\psi(h_{t-1}^1,x_{t-1}))$). 

The $\Big\{\mu_\psi\left(h_{t-1}^1,x_{t-1}\right)\Big\}_t$ and $\left\{\Sigma_\psi(h_{t-1}^1,x_{t-1})\right\}_t$ are parametrized by a network of GRU. 
This GRU is specifically designed to capture dependencies between individual steps. 
It consists of two fully connected layers, with one outputting an average vector and the other outputting a logarithmic vector of hidden variables.

\noindent\textbf{Posterior.}
For the posterior $q_\phi$ is to mimic of $p_\psi ($also $p)$.
It also uses the parameterised $p_\psi=p_\psi(h_{t<T}^1|u_{t<T},y_T).$
In the case of a reparameterisation using $p_\psi$ and a mean field decomposition, the posterior is given by Eq.4:
\begin{eqnarray}
  &q_\phi\left(h_{t<T}^1 \middle| x_{t<T},y_T\right)=\frac{q_\phi\left(y_T|h_{T-1}^1\right)}{q_\phi\left(y_T|x_{t<T}\right)}{\ast}\nonumber \\
  &\prod_{t<T}{q_\phi(h_t^1|x_t,h_{t-1}^1)},
\end{eqnarray}

\noindent The $q_\phi\left(h_t^1 \middle| x_t,h_{t-1}^1\right) \sim N\left(\mu\left(x_t,h_{t-1}^1\right), \Sigma\left(x_t,h_{t-1}^1\right)\right)$.

In particular, the posterior network $q_\phi\left(h_t^1 \middle| h_{t-1}^1,x_t\right)$ is used to encode input $x_t$ by a two-layer fully-connected layer.
\\
\\
\textbf{Multistep Hidden Markov Mechanism}

\noindent The Multistep Hidden Markov Model utilizes a neuralized Hidden Markov Model with an attention mechanism, consisting of a GRU with shared parameter $\theta_{2}$, an attention mechanism with shared parameter $\pi$ and and the latent variable $h_{t}^{i}(i=2, \dots, m)$.
The input to this mechanism is obtained by feeding the latent variables from the lower level, $u_{t}^{i-1} = \{h_{t-k+1}^{i-1}, h_{t-k+2}^{i-1}, \ldots, h_{t}^{i-1}\}$ (with time step k), into the attention mechanism. 
Subsequently, The GRU utilizes this input to generate the latest hidden variable.
Based on the Multistep Markov Condition \cite{pearl2009causality}, the corresponding joint distribution of the Hidden Markov chain is:
\begin{eqnarray}
  &p(h_{t<T}^i,h_{t<T}^{i-1},y_T) = p(y_{T+1} | h_T^i) {\ast} \nonumber \\
  &\prod_{t=1}^{\lfloor\frac{T}{k}\rfloor} p(h_{kt}^i | h_{k(t-1)}^{i},u_{kt}^{i-1})p(u_{kt}^{i-1}) | h_{kt}^i),
\end{eqnarray}
The $\{p(h_{kt}^i| h_{k(t-1)}^i,u_{kt}^{i-1})\}_t,q_\phi(h_{t<T}^i | h_{t<T}^{i-1},y_T),p(u_{kt}^{i-1} | h_{kt}^i)$ denote the a priori model, the posterior model and the generative model, respectively. 

\noindent
\textbf{Priori.} For the prior $p_\psi(h_{kt}^i | h_{k(t-1)}^i,u_{kt}^{i-1})$, it can be written as:
\begin{eqnarray}
  &p_\psi(h_{kt}^i | h_{kt-1}^i,u_{kt}^{i-1}) = \prod_{o} p_\psi(o_{kt}^{i} | o_{k(t-1)}^{i}, u_{kt}^{i-1}),
\end{eqnarray}
\frenchspacing
$\text{where } o \in h$, $p_\psi(h_{kt}^i | h_{k(t-1)}^i,u_{kt}^{i-1})$ for each $kt$ is distributed as $N(\mu_\psi(h_{k(t-1)}^i,u_{kt}^{i-1}),\Sigma _\psi(h_{k(t-1)}^i,u_{kt}^{i-1}))$.

$\text{The }\{\mu_\psi(h_{k(t-1)}^i, u_{kt}^{i-1})\}_t(\text{and}\{\Sigma_\psi(h_{k(t-1)}^i, u_{kt}^{i-1})\}_t)$\space are parametrized by a GRU.
An attention mechanism is employed to assign weights to different inputs at the input stage. This attention mechanism is specifically designed to capture multistep dependencies, allowing the model to effectively consider and prioritize relevant information from various sources.
After passing through the GRU, which enables the model to capture sequential dependencies within each input, there are two fully connected layers. after the GRU are two fully connected layers, one outputting the mean vector and one outputting the logarithmic vector of hidden variables.

\noindent
\textbf{Posterior.}
For the posterior $q_\phi$ is an closingly mimic for $p_\psi$(also $p$).
It also uses the heavily parameterised $p_\psi = p_\psi(h_{t<T}^i | h_{t<T}^{i-1},y_T)$. In the case of the reparameterisation using $p_\psi$ and the mean field decomposition, the posterior can be expressed by the following equation:

\begin{eqnarray}
  &q_\phi(h_{t<T}^{i-1} | h_{t<T}^{i-1},y_T) = \frac{q_\phi(y_T | h_{T-1}^{i})}{q_\phi(y_T | h_{t<T}^{i-1})}\ast\nonumber \\
  &\prod_{t <\left\lfloor\frac{T}{k}\right\rfloor} q_\phi(h_{kt}^{i} | h_{k(t-1)}^{i},u_{kt}^{i-1}), 
\end{eqnarray}

\noindent$\text{The }q_\phi(h_{kt}^i | h_{k(t-1)}^{i},u_{kt}^{i-1})\text{ follows a normal distribution with }\\ \text{mean } \mu(h_{k(t-1)}^i,u_{kt}^{i-1}) \ \text{ and covariance }\Sigma(h_{k(t-1)}^i,u_{kt}^{i-1})$.

Similarly, the posterior network represented by symbol $q_\phi(h_{kt}^i | h_{k(t-1)}^i, u_{kt}^{i-1})$ is encoded by a learnable weight matrix, weighting the input $u_{kt}^{i-1}=\{h_{k(t-1)+1}^{i-1},h_{k(t-1)+2}^{i-1},\ldots, \\h_{kt}^{i-1}\}$, and feeding it into a two-layer fully connected network.
\noindent\subsection{Learning Method}
\noindent To learn our proposed Pyramidal Hidden Markov Model, we first introduce a sequential VAE structure, with the network architecture shown in Fig. \ref{fig3}.
This ELBO is accompanied by a simplified $q_\emptyset(H_{1:T}|x_{1:T-1},y_T)$$(H=\{h^1,h^2,\ldots,h^m\})$ distribution as:\\
\begin{eqnarray}
  \mathbb{E}_{p(x_{t<T},y_T)}\left[L_{q_\emptyset,p_\psi}\right],
\end{eqnarray}

\noindent where $L_{q_\emptyset,p_\psi}=\left[\mathbb{E}_{q_\phi(H_{t<T}|x_{t<T},y_T)}log(\frac{p_\psi(H_{t<T},x_{t<T},y_T)}{q_\phi(H_{t<T}|x_{t<T},y_T)}) \right]$
\\

\noindent\textbf{Generated Part.}
For each time $t$, the generative model is a p.d.f. of a Gaussian distribution parameterised by a two-layer fully connected network to reconstruct the input sequence $x$. 
The $q_\phi(y_T|H_{T-1})$ is designed by a classifier with softmax as activation function or a predictor with no activation function.
\\

\noindent\textbf{Reformulation.}
Substituting the posterior $p_\psi$ and the prior $q_\emptyset$ in the above equation, we reformulate the ELBO as
\begin{eqnarray}
  \mathbb{E}_{p(x_{t<T},y_T)}\left[\log q_\emptyset\left(y_T\middle| x_{<T}\right)+\sum_{t<T}\mathcal{L}_{q_\emptyset,p_\psi}^t\right],
\end{eqnarray}
\begin{eqnarray}
  \mathcal{L}_{q_\emptyset,p_\psi}^t = \mathbb{E}_{p\left(H_t\middle| x_t,H_{t-1}\right)}\left[\log p_\psi\left(x_t\middle| H_t\right)\right]- \nonumber \\
  D_{KL}(q_\emptyset\left(H_t\middle| x_t,H_{t-1}\right)p_\psi\left(H_t\left|H_{t-1}|x_t\right)\right),
\end{eqnarray}
\begin{equation}
  \mathcal{L}_{q_\emptyset,p_\psi}^{T-1} = \mathbb{E}_{q_\emptyset}\left(H_{T-1}\middle| x_{T-1},H_{T-2}\right) [\ell_1 + \ell_2 + \ell_3].
\end{equation}
Here $\ell_1$, $\ell_2$, and $\ell_3$ are defined respectively:
\begin{align*}
  &\ell_1 \coloneqq \log(p_\psi (x_{T-1} \mid H_{T-1})), \\
  &\ell_2 \coloneqq \log \left(\frac{{p_\psi (y_T \mid H_{T-1})}}{{q_\emptyset (y_T \mid H_{T-1})}}\right), \\
  &\ell_3 \coloneqq \log \left(\frac{{p_\psi (H_{T-1} \mid H_{T-2}, x_{T-2})}}{{q_\emptyset (H_{T-1} \mid x_{T-1})}}\right).
\end{align*}

Since $p_\psi$ is an approximation to $q_\emptyset$, we parameterise $p_\psi\left(y_T\middle| H_{T-1}\right)$ as $q_\emptyset\left(y_T\middle| H_{T-1}\right)$ by reducing its $\ell_2$ degenerate to 0. Furthermore, we integrate $q_\emptyset\left(y_T\middle| x_{<T}\right)$ as follows:

\begin{eqnarray}
\int&\left(\prod_{t=1}^{T-1} q_\emptyset\left(H_t\middle| x_t,H_{t-1}\right)\right) q_\emptyset\left(y_T\middle| H_{T-1}\right) dH_1 \dots dH_{T-1}.\nonumber
\end{eqnarray}
\\
\textbf{Train and Test.} With such a reparameterisation, the reformulated ELBO in Eq.8 is our maximisation objective. In the inference phase, we obtain $H_t$ by iterating over the posterior network at each time step. Ultimately, we feed $H_t$ into the predictor $q_\phi(y_T|H_{T-1})$ to predict $y_T$.
\section{Experiment}
We evaluate the time series forecasting capability of Pyramidal Hidden Markov Model (PHMM). 
To have better understand the data that is suitable for our model, we train on datasets of varying size and evaluate. 
When considering the computational cost of train the model, PHMM performs very favourably, attaining state of the art on 3 Multivariate Time Series (MTS) classification benchmarks at a relative lower cost.
In addition, we present 8 SOTA time series classification approaches. 
Then, our model is compared with these 8 SOTA approaches across 20 UEA multivariate ties series datasets, and an ablation study is conducted using the neuralized Hidden Markov Model (HMM) as a baseline. 
Finally, we acquire a dataset of moderate size for the purpose of further examining the long-term predictive capacity of our model and also conducting an additional ablation analysis.

This section is concerned with testing the various properties of the model. 
Training and deployment of the network is conducted on a PC equipped with an Intel i9 CPU with a 32 GB mian memory and an NVIDIA GTX Geforce 2080ti GPU.
We then evaluate our model against train/test datasets provided by the dataset authors.
The preprocessing solely involves scaling without incorporating any data augmentation techniques.
\subsection{Dataset}
\begin{table*}[!ht]
  \caption{Classification results}
  \label{tab:classification result}
  \resizebox{\textwidth}{!}{%
  \begin{tabular}{c|c c c|c c c c c c c c|c c}
  \hline
  Dataset & \makecell{EDI} & \makecell{DTWI} & \makecell{DTWO} & \makecell{WEASEL\\+MUSE\\(2018)} & \makecell{MLSTM\\-FCN\\(2019)} & \makecell{MrSEQL\\(2019)} & \makecell{TapNet\\(2020)} & \makecell{ShapeNet\\(2021)} & \makecell{ROCKET\\(2020)} & \makecell{MiniRocket\\(2021)} & \makecell{RLPAM\\(2022)} & \makecell{HMM} & \makecell{PHMM\\(ours)} \\
  \hline
  \hline
  IW & 0.127 & N/A & 0.116 & N/A & 0.168 & N/A & 0.210 & 0.251 & N/A & 0.594 & 0.351 & 0.216 & \textbf{0.657} \\
  FD & 0.521 & 0.514 & 0.531 & 0.546 & 0.545 & 0.545 & 0.557 & 0.603 & 0.626 & 0.622 & 0.622 & 0.521 & \textbf{0.630} \\
  FM & 0.530 & 0.520 & 0.550 & 0.490 & 0.560 & 0.550 & 0.530 & 0.570 & 0.530 & 0.530 & 0.610 & 0.590 & \textbf{0.640} \\
  LSST & 0.457 & 0.574 & 0.552 & 0.592 & 0.371 & 0.589 & 0.567 & 0.589 & 0.638 & 0.645 & 0.646 & 0.550 & \textbf{0.722} \\
  SRS1 & 0.771 & 0.764 & 0.776 & 0.711 & 0.875 & 0.680 & 0.653 & 0.781 & 0.847 & 0.922 & 0.802 & 0.772 & \textbf{0.925} \\
  SRS2 & 0.483 & 0.533 & 0.539 & 0.460 & 0.472 & 0.572 & 0.550 & 0.530 & 0.578 & 0.540 & 0.632 & 0.622 & \textbf{0.640} \\
  AF & 0.267 & 0.267 & 0.200 & 0.333 & 0.260 & 0.267 & 0.33 & 0.40 & 0.067 & 0.133 & 0.733 & 0.467 & \textbf{0.733} \\
  BM & 0.675 & 1.000 & 0.975 & 1.000 & 0.660 & 0.950 & 0.75 & 0.75 & \textbf{1.000} & \textbf{1.000} & \textbf{1.000} & 0.737 & \textbf{1.000} \\
  CR & 0.944 & 0.986 & \textbf{1.000} & \textbf{1.000} & 0.917 & 0.986 & 0.958 & 0.986 & \textbf{1.000} & 0.986 & 0.764 & \textbf{1.000} & \textbf{1.000} \\
  JV & 0.925 & 0.960 & 0.950 & 0.974 & 0.975 & 0.921 & 0.966 & 0.985 & 0.964 & 0.980 & 0.935 & 0.743 & \textbf{0.989} \\
  HB & 0.622 & 0.660 & 0.718 & 0.725 & 0.662 & 0.744 & 0.752 & 0.757 & 0.726 & 0.772 & \textbf{0.780} & 0.738 & \textbf{0.780} \\
  MI & 0.510 & 0.390 & 0.500 & 0.500 & 0.510 & 0.520 & 0.590 & 0.610 & 0.560 & 0.55 & \textbf{0.610} & 0.570 & \textbf{0.610} \\
  CT & 0.965 & 0.965 & 0.970 & 0.965 & 0.890 & 0.970 & 0.89 & 0.90 & N/A & 0.965 & \textbf{0.978} & 0.816 & 0.970 \\
  AWR & 0.969 & 0.978 & 0.986 & 0.989 & 0.512 & 0.991 & 0.590 & 0.611 & \textbf{0.992} & 0.991 & 0.922 & 0.898 & 0.970 \\
  EP & 0.667 & 0.977 & 0.965 & \textbf{1.000} & 0.762 & 0.993 & 0.981 & 0.987 & 0.987 & \textbf{1.000} & 0.827 & 0.972 & 0.980 \\
  EC & 0.295 & 0.306 & 0.325 & 0.432 & 0.375 & \textbf{0.557} & 0.325 & 0.317 & 0.376 & 0.468 & 0.358 & 0.356 & 0.452 \\
  HMD & 0.230 & 0.307 & 0.232 & 0.366 & 0.363 & 0.150 & 0.380 & 0.342 & 0.447 & 0.391 & \textbf{0.636} & 0.432 & 0.514 \\
  LIB & 0.834 & 0.884 & 0.862 & 0.878 & 0.846 & 0.862 & 0.840 & 0.857 & 0.908 & \textbf{0.925} & 0.796 & 0.551 & 0.879 \\
  RS & 0.867 & 0.843 & 0.813 & \textbf{0.923} & 0.812 & 0.859 & 0.859 & 0.872 & 0.920 & 0.858 & 0.858 & 0.743 & 0.842 \\
  UWGL & 0.883 & 0.870 & 0.902 & 0.935 & 0.892 & 0.873 & 0.896 & 0.902 & 0.937 & 0.937 & \textbf{0.945} & 0.418 & 0.916 \\
  \hline
  Avg. Rank &8.350&7.250&6.900&5.350&7.600&5.950&6.100&4.950&4.300&3.400&4.050&7.150&\textbf{2.300}\\
  MPCE &0.037&0.030&0.033&0.027&0.038&0.029&0.034&0.032&0.027&0.026&0.026&0.039&\textbf{0.021}\\
  Win\textbackslash Ties &0&1&1&4&0&1&0&1&3&3&7&1&\textbf{12}\\
  Our 1-to-1-Wins &19&16&17&14&20&15&18&17&13&13&12&19&-\\
  Our 1-to-1-Losses &1&3&1&4&0&4&2&2&5&6&4&0&-\\
  Wilcoxon Test p-value &0.000&0.000&0.000&0.011&0.000&0.001&0.000&0.011&0.113&0.321&0.000&0.000&-\\
  \hline
  \end{tabular}%
  }
  \begin{spacing}{0.6}
  \resizebox{\textwidth}{!}{-Accuracy results are sorted by deviations between PHMM and the best performing baselines. The classification accuracy of the baselines on the UEA archive datasets are obtained}

  \resizebox{\textwidth/100*50}{!}{from their original papers, except ROCKET which is run using the code open-sourced \cite{DBLP:journals/corr/abs-2102-05765}.}
\end{spacing}
\end{table*}
We first evaluate our model on 20 datasets from UEA MTS classification archive \cite{dauUCRTimeSeries2019, UEA2019},
which consists of data collected from various domains, including Articulary Word Recognition (AWR), Atrial Fibrillation (AF), Basic Motion (BM), etc.
The datasets contain a varying number of samples, ranging from 27 to 50,000, with sample lengths ranging from 8 to 17,901 \cite{UEA2019}. 

Furthermore, we collecte real-world stock data with the aim of making long-term predictions. 
These data are obtained from publicly available daily stock prices in the market. 
Subsequently, we processe and standardized the gathered dataset, which comprise a total of 4732 stocks. 
Among these, 3312 stocks are utilized as the training dataset while the remaining 1420 stocks serve as the test dataset. 
The sample points on each stock are fixed at 200, and our prediction model utilizes the initial 160 sample points to forecast the subsequent 40 sample points.
We apply our model to this Stocks dataset and compared it with other methods. 
As referenced in \cite{karunasingha2022root}, the performance assessment of our model is criterion-led by the Root Mean Square Error (RMSE), with the RMSE value of the LSTM model serving as our baseline.
\subsection{Evaluation Criteria}
In the experimental session, we present various performance metrics to evaluate the effectiveness of our classification model on different datasets. 
As referenced in \cite{gao2022reinforcement}, these metrics include classification accuracy, average ranking, and wins/losses for each dataset when tested with our model. 

To further assess the performance of our model across all datasets, we compute the Mean Per Class Error (MPCE). This metric provides an average measure of error per class by considering all datasets collectively. By applying the described procedure as depicted in Eq.\ref{eq1}, we calculate the MPCE for each dataset individually and then aggregate them to obtain an overall assessment.
\begin{eqnarray}
  \eta = \frac{1}{n} \sum_{i=1}^{n} |v_i| = \frac{1}{n} \sum_{i=1}^{n} |x_i - \bar{x}|.
  \label{eq1}
\end{eqnarray}

We also performed the Friedman test and the Wilcoxon signed rank test using Holm's $\alpha$ (5\%) following the procedure described therein \cite{zimmerman1993relative}.
The Friedman test \cite{sheldon1996use} is a non-parametric statistical test used to demonstrate the significance of differences in performance across all methods.
The Wilcoxon signed rank test \cite{woolson2007wilcoxon} is a non-parametric statistical test based on the assumption that the median of the ranks is the same between our methods and any baseline. 

\subsection{Baseline}
In our benchmarks datasets, we compared the purposed Pyramidal Hidden Markov Model (PHMM) with SOTA methods to highlight the effectiveness of the purposed PHMM.
In those experiments, the methods are trained independently.
Those baselines include:
\noindent
\begin{enumerate}[0]
  \item [$\bullet$] \textit{HMM \cite{eddy1996hidden, Mehta_2022}.} Directly using UCSs as inputs to the traditional HMM denoted as HMM. This serves as an ablation baseline.
  \item [$\bullet$] \textit{WEASEL+MUSE \cite{schafer2017multivariate}.} A sliding-window method to create a feature vector for each dimension of the time series, extracting unique features for each dimension.
  \item [$\bullet$] \textit{MLSTM-FCN \cite{karim2019multivariate}.} The utilization of a fusion between multi-layer recurrent neural networks and full convolutional networks.
  \item [$\bullet$] \textit{MrSEQL \cite{nguyen2021mrsqm}.} The MrSEQL employs symbolic representations of Symbolic Aggregate Approximation (SAX) and Symbolic Feature Analysis (SFA) to capture diverse characteristics of time series data.
  \item [$\bullet$] \textit{TapNet \cite{zhang2020tapnet}.} A new network that trains feature representation by considering proximity to class prototypes, even with limited data labels.
  \item [$\bullet$] \textit{ShapeNet \cite{li2021shapenet}.} The shapelet candidates of different lengths using a neural network and generates representative final shapelets instead of using all embeddings directly.
  \item [$\bullet$] \textit{ROCKET \cite{dempster2020rocket}.} The Rocket uses a simple linear classifier of random convolutional kernels instead of convolutional neural networks, and combines a small number of existing methods to achieve superior performance.
  \item [$\bullet$] \textit{MiniRocket \cite{dempster2021minirocket}.} A powerful variant of the ROCKET model.
  \item [$\bullet$] \textit{RLPAM \cite{gao2022reinforcement}.} The RLPAM employs a reinforcement learning (RL) guided PAttern Mining framework (RLPAM) to discern meaningful and significant patterns for the classification of MTS.
  \item [$\bullet$] \textit{EDI, DTWI and DTWO \cite{dulac1512deep, bagnall2018uea}.} EDI is a nearest neighbour classifier based on Euclidean distance. DTWI is dimension-independent DTW. DTWD is dimension-dependent DTW. 
\end{enumerate}

\subsection{Performance on 20 UEA Datasets}
The overall 20 UEA datasets' accuracy results are shown in Table \ref{tab:classification result}. 
The result `N/A' indicates that the corresponding method was not demonstrated or could not produce the corresponding result. 
Overall, PHMM achieves optimality among all compared methods. 
In particular, PHMM obtained an average ranking of 2.300, which is above all baselines.
  PHMM resulted in 12 win/ties, while the best SOTA result was 7 win/ties.
  In the mentioned MPCE, PHMM achieves the lowest error rate of all datasets.
  In the Friedman test, our statistical significance is $p\leq0.001$. This demonstrates the significant difference in performance between this method and the other seven methods. 
  A Wilcoxon signed rank test is performed between PHMM and all baselines, which showed that PHMM outperformed the baseline on all 20 UEA datasets at a statistical significance level of $p \leq 0.05$, except for ROCKET, MiniRocket and RLPAM. 
  Interestingly, at UCS performance is quite good but not as good as the SOTA. 
  On most datasets, PHMM can significantly outperform traditional HMM, even better than the SOTA. 
  On all datasets where HMM performs poorly, PHMM can significantly improve its performance, but may not beat the SOTA on all datasets. 
  The above observations demonstrate the importance of Multistep Stochastic states in for MTS classification and prediction problems, as proposed in PHMM. 
  Moreover, PHMM performs better on datasets with a limited number of training samples, such as AtrialFibrillation (AF), which contain only 15 training samples each. 
  The reason may be that these datasets inherently contain high-quality representative and discriminative patterns captured by PHMM.
\subsection{Performance on the Stocks Dataset}
\begin{table}[ht]
  \vspace{-0.5cm}
  \centering
  \caption{Prediction result}
  \label{tab:forecast result}
  \resizebox{\columnwidth/10*8}{!}{
    \begin{tabular}{c|c c c c}
      \toprule
      \makecell{Method} & \makecell{LSTM} & \makecell{ND} & \makecell{HMM}  & \makecell{PHMM} \\
      \hline                                              
      \makecell{RMSE} & \makecell{5.856} & \makecell{6.039}&  \makecell{7.857} & \makecell{\textbf{2.638}} \\
      \makecell{Ratio} & \makecell{1.000} & \makecell{1.031} &  \makecell{1.341}  & \makecell{\textbf{0.450}} \\
      \bottomrule
    \end{tabular}%
  }
\end{table}  
\noindent We further investigate the effective utilization of PHHM in real-world long-term forecasting tasks. 
The comprehensive RMSE results for Stocks are presented in Table \ref{tab:forecast result}, obtained through replicated experiments with a significance level of $p \leq 0.1$. 
ND is a method of fitting data based on relevant mathematical functions.
We employ RMSE as a metric for assessing the performance of each model, where a lower RMSE value is considered superior to a higher one.
The Ratio is calculated using LSTM's RMSE as a baseline. 
In particular, PHMM obtain an lowest RMSE of 2.638, which is under all baselines.
Overall, PHMM outperforms all other compared methods and demonstrates optimality. 

Notably, PHHM demonstrates a superior performance compared to the conventional HMM approach in terms of long-term forecasting accuracy.
In the Friedman test, our statistical significance is $p\leq0.001$. This demonstrates the significant difference in performance between this method and HMM. 
In particular, for PHMM, PHMM performs much better than HMM when the step size and stacking parameters of the model are well-designed.
It can be seen that the PHMM has the ability of long-term dependence capture and is more robust to non-stationary time series (e.g., stock market, etc.) than HMM.

\subsection{A Case for Tunable Hyper-parameters}
In this section, we further investigate the Tunable hyper-parameters' impact on the performance of our model. 
The first hyper-parameter under consideration is timesteps $k$, and the second is the number of stacked layers $m$.
We conduct our experiments using the AWR dataset, where the series length of the dataset is 144.
\begin{table}[ht]
  \vspace{-0.5cm}
  \centering
  \caption{Different tunable hyper-parameters results}
  \label{tab:tunable result}
  \resizebox{\columnwidth}{!}{
    \begin{tabular}{c|c c c c c c}
    \toprule
      \makecell{\diagbox [width=7.5em,trim=l] {Number\\of layers}{Timesteps}} & \makecell{$k = 1$} & \makecell{$k = 3$}  & \makecell{$k = 5$} & \makecell{$k = 7$} & \makecell{$k = 9$} & \makecell{$k = 11$}\\
      \hline                                              
      \makecell{$m=2$} & \makecell{0.910} & \makecell{0.910}& \makecell{0.920} &  \makecell{0.943} &  \makecell{0.953}& \makecell{0.957}\\
      \makecell{$m=3$} & \makecell{0.920} & \makecell{0.933}& \makecell{0.956} &  \makecell{0.960} &  \makecell{0.963}& \makecell{0.963} \\
      \makecell{$m=4$} & \makecell{0.923} & \makecell{0.940}& \makecell{0.960} &  \makecell{0.970} &  \makecell{0.963}& \makecell{0.963} \\
      \makecell{$m=5$} & \makecell{0.933} & \makecell{0.940}& \makecell{0.960} &  \makecell{0.970} &  \makecell{0.963}& \makecell{0.963} \\
      \makecell{$m=5$} & \makecell{0.943} & \makecell{0.943}& \makecell{0.960} &  \makecell{0.970} &  \makecell{0.963}& \makecell{0.963}\\
      
      \bottomrule
    \end{tabular}%
  }
\end{table}  
\\

As the accuracy results are shown in Table \ref{tab:tunable result},
the parameters play a crucial role in determining the performance of our model. It is essential to carefully select appropriate timesteps and the number of stacked layers to ensure optimal results. The choice of timesteps should strike a balance between capturing sufficient temporal information and avoiding excessive computational complexity.
Additionally, the depth of PHHH should be moderate. A shallow architecture may not capture complex patterns effectively, while an excessively deep one can lead to unnecessary prolongation of training time. Striking the right balance is pivotal in attaining precise predictions.
\section{Conclusions}
In this study, we propose a Pyramid Hidden Markov Model (PHMM) composed of an input hidden Markov mechanism and multi-step hidden Markov mechanisms. The input mechanism resembles a traditional HMM and captures fundamental stochastic states, while the multi-step mechanism is capable of capturing short stochastic states. By integrating these two mechanisms, the PHMM establishes long-term dependencies, leading to accurate and enhanced forecasting. Comprehensive experiments conducted on several multivariate time series datasets demonstrate that our model surpasses its competitive counterparts in time series forecasting.
\label{sec:illust}
\bibliographystyle{IEEEbib}
\bibliography{main}

\end{document}